%% file: main.tex
\documentclass[10pt,twocolumn,letterpaper]{article}

\usepackage[pagenumbers]{cvpr} %

\definecolor{cvprblue}{rgb}{0.21,0.49,0.74}
\usepackage[pagebackref,breaklinks,colorlinks,allcolors=cvprblue]{hyperref}

\usepackage{xspace}
\usepackage{caption, overpic, textpos}

\usepackage{amssymb}
\usepackage{wrapfig}
\usepackage{multirow}
\usepackage{enumitem}

\usepackage[linesnumbered, ruled]{algorithm2e}
\usepackage{algpseudocode}%
\usepackage{bm}
\usepackage{array}
\usepackage{verbatim}
\usepackage{algorithmicx}
\usepackage{comment}
\usepackage{lipsum}
\usepackage{stringstrings}
\usepackage{makecell}
\usepackage{url}
\usepackage{amsmath}
\usepackage{bbding}
\usepackage{pifont}
\usepackage{wasysym}
\usepackage{xcolor}
\usepackage{makecell}

\usepackage[capitalize]{cleveref}
\crefname{section}{Sec.}{Secs.}
\Crefname{section}{Section}{Sections}
\Crefname{table}{Table}{Tables}
\crefname{table}{Tab.}{Tabs.}

\newcommand{\myparagraph}[1]{{\setlength{\parskip}{0.3em} \noindent \textbf {#1}}}

\newcommand{\xinyu}[1]{{\color{black}#1}}

\definecolor{darkgreen}{rgb}{0.0, 0.5, 0.0} %

\makeatletter  \newcommand\figcaption{\def\@captype{figure}\caption}
\newcommand\tabcaption{\def\@captype{table}\caption}

\title{
Training-Free Motion-Guided Video Generation with Enhanced Temporal Consistency Using Motion Consistency Loss}

\author{
Xinyu Zhang\textsuperscript{\rm 1}\quad
Zicheng Duan\textsuperscript{\rm 1}\quad
Dong Gong\textsuperscript{\rm 2}\quad
Lingqiao Liu\textsuperscript{\rm 1}\\[0.5em]
\small
\textsuperscript{\rm 1}The University of Adelaide\quad
\textsuperscript{\rm 2}The University of New South Wales\\[0.5em]
{\tt\small\url{https://zhangxinyu-xyz.github.io/SimulateMotion.github.io/}}
}

\begin{document}
\maketitle
\input{sec/0_abstract}    
\input{sec/1_intro}

\input{sec/2_related_work}

\input{sec/3_method}

\input{sec/4_experiment}

\input{sec/5_conclusion}

{
    \small
    \bibliographystyle{ieeenat_fullname}
    \bibliography{main}
}

\input{sec/X_suppl}

\end{document}

%% file: sec/0_abstract.tex
\begin{abstract}
In this paper, we address the challenge of generating temporally consistent videos with motion guidance. While many existing methods depend on additional control modules or inference-time fine-tuning, recent studies suggest that effective motion guidance is achievable without altering the model architecture or requiring extra training. Such approaches offer promising compatibility with various video generation foundation models. However, existing training-free methods often struggle to maintain consistent temporal coherence across frames or to follow guided motion accurately.
In this work, we propose a simple yet effective solution that combines an initial-noise-based approach with a novel motion consistency loss, the latter being our key innovation. Specifically, we capture the inter-frame feature correlation patterns of intermediate features from a video diffusion model to represent the motion pattern of the reference video. We then design a motion consistency loss to maintain similar feature correlation patterns in the generated video, using the gradient of this loss in the latent space to guide the generation process for precise motion control. This approach improves temporal consistency across various motion control tasks while preserving the benefits of a training-free setup. Extensive experiments show that our method sets a new standard for efficient, temporally coherent video generation.

\end{abstract}

%% file: sec/1_intro.tex
\section{Introduction}
\label{sec:intro}
Recent advancements in video diffusion models~\cite{ho2022imagen, singer2022make, he2022lvdm, wang2023modelscope,zeroscope,chen2023videocrafter1,chen2024videocrafter2,svd,hong2022cogvideo,yang2024cogvideox}, have greatly improved the quality of videos generated from text instructions. Nonetheless, in many cases, text instructions alone may fall short of fully conveying the user’s intent, emphasizing the need for additional guidance in video generation. Motion customization presents a promising solution by enabling the generation of videos that integrate content from text descriptions with precise motion guidance drawn from a reference movement. This approach proves  valuable in many scenarios. For instance, it can be useful for creating content that is unlikely to occur in real-world scenarios, such as ``a panda lifting weights''.  Through motion customization, it becomes possible to adapt the ``lifting weights'' motion style from a ``human lifting weights'' video to animate a panda.

\begin{figure}[t]
\centering
\includegraphics[trim =0mm 0mm 0mm 0mm, clip, width=1.0\linewidth]{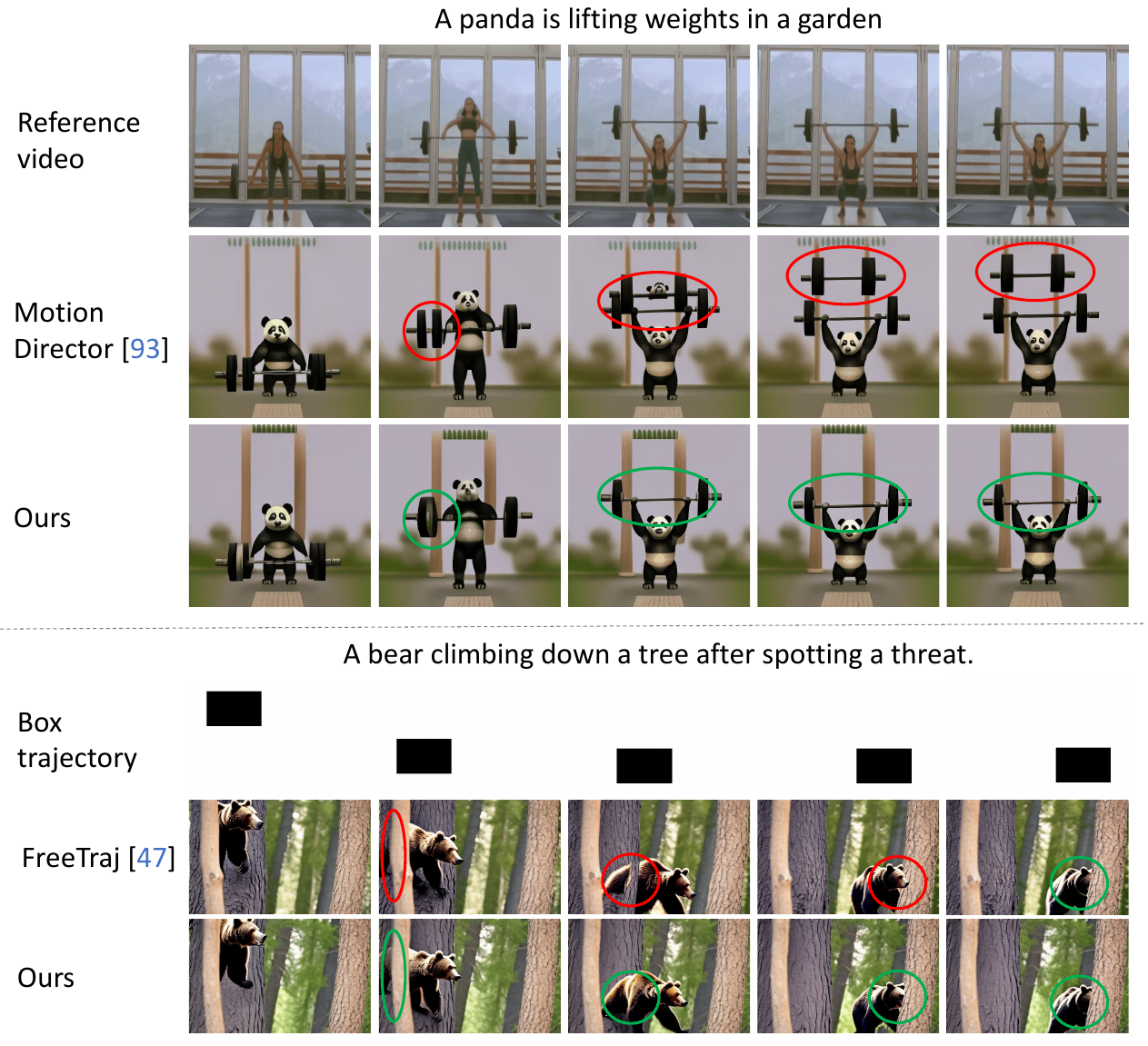}
\caption{Visualization comparisons on our method and two existing motion customization methods, including the reference video based Motiondirector~\cite{zhao2025motiondirector}, and the bounding box trajectory based FreeTraj~\cite{qiu2024freetraj}.
Methods in the upper part use the inversion noise from the reference video, while methods in the lower part use the well-designed noise as initialization.
The \textcolor{red}{red} circle regions represent the inconsistent temporal coherent, while the \textcolor{darkgreen}{green} circle regions represent the correct one.
}
\vspace{-0.5cm}
\label{fig:figure1}
\end{figure}

Existing approaches to motion customization take various guidance types, such as reference videos, trajectories, bounding box movements, or motion fields, as control signals. Many of these methods depend on additional modules, including Low-Rank Adaptations (LoRAs)~\cite{hu2021lora, zhao2025motiondirector, ren2024customize, wu2024customcrafter}, adapters or embeddings~\cite{liu2024video, wang2024motion, zhang2023adding, zhang2024tora, wei2024dreamvideo, wang2024motionctrl, wu2025draganything, niu2024mofa, li2024animate, li2024director3d}, or fine-tuning the video generation modules~\cite{zhang2023motioncrafter, chen2023motion, jeong2024vmc, wu2023tune, yang2024direct, shin2024edit, wu2024lamp}, which incur extra costs during either training or testing. When new foundation models or updates are released, these methods require retraining to adapt to the updated models.

This challenge has led to the development of training-free solutions~\cite{xiao2024video, trailblazer, qiu2024freetraj, jain2024peekaboo, jeong2025dreammotion, qiu2023freenoise, wu2023freeinit}, which aim to achieve motion guidance by designing specific noise initialization schemes~\cite{qiu2023freenoise, wu2023freeinit, qiu2024freetraj} and using attention mask operations~\cite{qiu2024freetraj, jain2024peekaboo, jeong2025dreammotion, xiao2024video}. Despite these advancements, many current methods still struggle with maintaining temporal coherence across frames. The lower part of Figure~\ref{fig:figure1} illustrates this issue, showing that existing methods often suffer from temporal inconsistency, \eg, the loss of detail on a bear when occluded by a tree.

To address the challenge of inconsistent temporal coherence, we propose a simple yet effective training-free solution in this paper. Our key insight is that better motion guidance can be achieved by combining two complementary approaches: implicit motion guidance through initial noise and explicit guidance via a motion pattern consistency loss.
Specifically, we begin by using the inversion noise from the reference video as the noise initialization for the video diffusion model. Recent studies~\cite{liu2024video, wu2023freeinit, qiu2024freetraj, zhao2025motiondirector, shin2024edit} have demonstrated the effectiveness of this strategy, as the initial noise serves as a strong prior that captures the movement within the video. However, our empirical findings indicate that relying solely on initial noise is not always effective; in some cases, it can lead to inconsistencies in appearance across frames, such as the case shown in 
Figure~\ref{fig:figure1}.

As a remedy, we propose a more explicit control scheme to enhance temporal consistency. We start by characterizing the motion pattern of the reference video through the feature correlation patterns from a diffusion model across different frames. We then design a loss function to encourage similar correlation patterns in the generated video, incorporating the gradient of this loss into the denoising inference process in a classifier-guidance-like manner~\cite{ho2022classifierfree}.

Moreover, rather than examining correlation between all pairs of local features across frames, we focus on the correlation patterns between a few key points in one frame and features in neighboring frames. This selective approach avoids overly rigid motion transfer, which is especially useful when the reference video motion does not perfectly align with the desired motion in the generated video.
Since the initial noise from the reference video already provides a reasonable level of motion control, and video diffusion models naturally maintain a degree of temporal consistency, we find that focusing on a few key points is sufficient to achieve strong temporal coherence.

Quantitative and qualitative experiments on three benchmarks, including LOVEU-TGVE-2023~\cite{wu2023cvpr}, UCF Sports Action~\cite{soomro2015action}, and prompts from \cite{qiu2024freetraj}, and real collected videos from Phone, show that our method can follow the motion movement with enhanced temporal coherent.
Our contributions are as follows:
\begin{itemize}[left=0em]
\item[$\bullet$] We propose a simple and effective training-free technique to improve the motion-guided video generation by enhancing frame-to-frame coherence.
\item[$\bullet$] We design an approach that represents the reference motion through the inter-frame feature correlation patterns of sparse points and transfers motion by replicating these reference matching patterns.
\item[$\bullet$] We demonstrate that the proposed approach complements the initial-noise-based motion guidance method. Together, they consistently enhance temporal coherence across a wide range of diverse and complex motions.
\end{itemize}

%% file: sec/2_related_work.tex
\section{Related Work}
\label{sec:related_work}

\myparagraph{Video diffusion models.}
A range of approaches approaches have been explored to achieve high-quality video generation, including Generative Adversarial Networks (GANs) \citep{vondrick2016generating, saito2017temporal, tulyakov2018mocogan, balaji2019conditional, tian2020good, shen2023mostgan}, autoregressive models \citep{srivastava2015unsupervised, yan2021videogpt, le2021ccvs, hong2022cogvideo, ge2022long, wang2024emu3, tian2024visual,kondratyuk2023videopoet}, and implicit neural representations \citep{yu2021generating, skorokhodov2021stylegan}.
With diffusion models~\cite{sdxl,svd,lvdm,rombach2021highresolution} appear, recent diffusion-based video generation models~\cite{ni2023conditional, yu2023video, mei2023vidm, voleti2022mcvd,ho2022imagen, singer2022make, he2022lvdm, VideoFusion, blattmann2023align, zhang2023show1, Wang2023LAVIEHV, guo2023animatediff, wu2023tune, khachatryan2023text2video} have shown high ability to produce high-quality generation.
These foundation models are usually pre-trained on large-scale video datasets \citep{chen2024panda, wang2023internvid, nan2024openvid, webvid10M}.
Some open-sourced foundation models, such as \cite{wang2023modelscope, zeroscope, hong2022cogvideo, yang2024cogvideox}, provide base video generation models for users to customize and build specific models.

\myparagraph{Controllable video generation.}
Controllable video generation aims to generate synthetic data according to the given explicit control signals, such as motion, human pose, layout, optical flows, etc. \citep{zhang2023adding, zhao2023uni, ma2023follow, geng2024motion}.
The controllable text-to-video generation methods focus on generating videos based on the control signals.
Some methods, \eg, VideoCrafter \citep{he2022lvdm}, VideoComposer \citep{wang2023videocomposer}, Control-A-Video \citep{chen2023control}, employ depths, sketches, or movement information from the reference videos as conditions to control motions, whose structures are like ControlNet~\cite{zhang2023adding}.
Many methods~\cite{zhao2025motiondirector, ren2024customize, wu2024customcrafter} have attempted to train lightweight modules, like Low-Rank Adaptations (LoRAs)~\cite{hu2021lora}, for motion customization to transfer the motion from the reference videos to the target ones.
However, these reference video based methods usually need to train different modules for different motion types, limiting the generalization ability.
Some approaches~\cite{chen2023motion,deng2025dragvideo,wang2024boximator,yang2024direct,huang2023fine,qiu2024freetraj,jain2024peekaboo,trailblazer} explore to use trajectories or bounding boxes to provide motion information, which are more flexible and user-friendly.

\begin{figure*}[t!]
\centering
\includegraphics[trim =0mm 0mm 0mm 0mm, clip, width=0.98\linewidth]{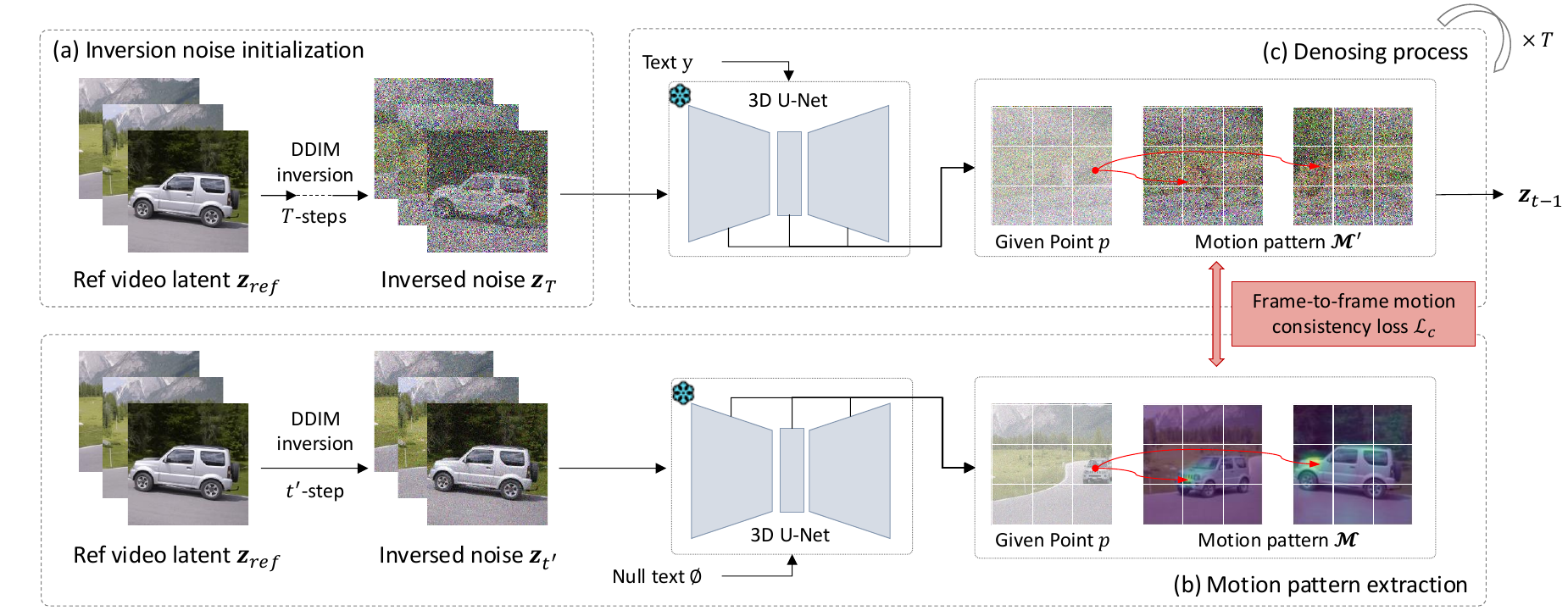}
\vspace{-0.3cm}
\caption{Overview of our method. We first conduct (a) inversion noise initialization on the reference video to obtain the initial noise $\mathbf{z}_T$ (Section~\ref{sec:initial_noise}). Then we (b) extract the motion pattern $\pmb{\mathcal{M}}$ from the reference video for each tracked point $p$ (Section~\ref{sec:motion_traj_extract}).
During the (c) denoising process, we use the proposed frame-to-frame motion consistency loss $\mathcal{L}_c$, calculated with Eq.~\ref{eq:consistency_loss} based on $\pmb{\mathcal{M}}$ and newly extracted $\pmb{\mathcal{M}}'$ from the noise $\mathbf{z}_t$ as the motion guidance for the noise estimation (Section~\ref{sec:consistency_loss}).
The detail of our method is in Algorithm~\ref{algo}.
}
\vspace{-0.3cm}
\label{fig:overview}
\end{figure*}

In this work, we focus on a general training-free approach, which can be integrated into various video generation models using various motion formats, \eg, reference video based or trajectory based motions, without requiring additional modules or training.

%% file: sec/3_method.tex
\section{Method}
\label{sec:method}

\subsection{Preliminaries}

\myparagraph{Video diffusion model.}
Video diffusion models generate videos by repeatedly denoising a randomly sampled sequence of Gaussian noises.
They are commonly represented as functions $\epsilon_\theta(\mathbf{z}_t, t, y)$. A common implementation of this mapping function is the 3D U-Net, which consists of down-sampling, middle, and up-sampling blocks. Each block includes multiple convolutional layers, as well as spatial and temporal transformers.
The function $\epsilon_\theta(\mathbf{z}_t, t, y)$ predicts the noise from a noisy latent $\mathbf{z}_t$ conditioned on $y$ at step $t$ of the diffusion process.
$\epsilon_\theta(\mathbf{z}_t, t, y)$ can be trained by minimizing:
\begin{equation}
    \mathcal{L} = \mathbb{E}_{\mathbf{z}_0, y, \epsilon\sim\mathcal{N}(0, \mathbf{I}), t\sim\mathcal{U}(0, T)}\left\lbrack 
\lVert \epsilon - \epsilon_\theta(\mathbf{z}_t, t, y) \rVert_2^2 \right\rbrack,
\label{eq:basic_train_loss}
\end{equation}
where $\mathbf{z}_0 \in \mathbb{R}^{4 \times F \times H \times W}$ is the latent code from the encoder, where $F$, $H$, $W$ represents the video length, height and width of the latent code.
$t\sim\mathcal{U}(0, T)$ is random sampled when training, where $\mathit{T}$ is the maximum timestep.

\myparagraph{Guidance in diffusion models.}
Classifier guidance is a widely used technique for injecting external control in content generation \cite{ho2022classifierfree,song2021scorebased,dhariwal2021diffusion,bansal2023ugd,geng2024motion}. It introduces a loss term, $\mathcal{L}_e$, to measure how well the generated latent aligns with the control target, integrating the gradient of this loss function into the denoising process to achieve controlled generation. Formally, it augments the noise estimate with the following formula:
\begin{equation}
    \label{eq:guidance}
    \hat{\epsilon}_\theta({\mathbf z}_t, t, y) := \epsilon_{\theta}(\mathbf{z}_t, t, y) + \sigma_t \nabla_{\mathbf{z}_t} \mathcal{L}_e(\mathbf{z}_t),
\end{equation}
where, $\hat{\epsilon}_\theta$ represents the modified noise estimate, and $\sigma_t$ is a weighting schedule. These methods offer the advantage of not requiring model retraining, as they simply update the score function calculation using the gradients of $\mathcal{L}_e$.

\subsection{Inversion Noise Initialization}
\label{sec:initial_noise}
Several image and video editing studies~\cite{hertz2022prompt, qiu2023freenoise, wu2023freeinit, jeong2025dreammotion,qiu2024freetraj,shin2024edit} have demonstrated that initial noises significantly impact the generated content. For example, some reference video-based or trajectory-based video generation methods~\cite{shin2024edit, qiu2024freetraj, qiu2023freenoise} have shown that the direction of motion in the generated video often aligns with the motion flow present in the noise initialization. Building on this concept, recent works like \cite{qiu2024freetraj} inject a user-provided motion trajectory into customized initial noise, enabling motion control within the video. In another approach, \cite{shin2024edit} optimizes a null-text embedding to capture and replicate the motion in a reference video.

Our method is grounded in the same principle: we treat a reference video—either provided directly by the user or synthesized based on a user-defined trajectory—as the primary guide for motion in the generated video. To incorporate this motion information, we employ DDIM~\cite{ddim} inversion to convert the reference video into a specific Gaussian noise, denoted as $\mathbf{z}_T$. This inversion process captures the motion characteristics of the reference, embedding them as initial noise. Once this initial noise is established, one can use an existing video diffusion model to generates a video based on this motion-guided initialization.

As illustrated in the third row in Figure~\ref{fig:ablation_compar}, while this straightforward approach may introduce some artifacts in the generated video, it still manages to follow the reference video’s motion to a significant extent. This indicates that the initial noise derived from reference video inversion can effectively guide the generated video’s movement, preserving key aspects of the motion while allowing room for refinement.

\subsection{Motion Pattern Extraction}
\label{sec:motion_traj_extract}

As discussed in the Introduction (Section~\ref{sec:intro}), this paper aims to establish a more explicit motion control to enhance the temporal coherence of initial-noise-based approaches. The first question we address is how to effectively represent motion patterns. Traditionally, dense motion fields~\cite{wu2025draganything,chen2023control,niu2024mofa} or optical flow~\cite{geng2024motion,dragnvwa} have been used for this purpose. However, these are calculated at the pixel level, making it inconvenient for guiding video generation, as one would need to generate the video first and then verify if the motion pattern aligns with the reference video.

In this paper, we propose an alternative solution by using inter-frame feature correlation patterns to represent the motion pattern of the reference video. This is a reasonable assumption, as the features from a diffusion model can capture certain semantic correspondence \cite{hedlin2024unsupervised,nam2024dreammatcher}. Specifically, %
\textcolor{black}{we first extract the feature maps $\{\mathbf{F}_l\}_{l=1}^{L}$ from all temporal attention modules in the video diffusion model with the input of $\mathbf{z}_{t'}$, where $\mathbf{z}_{t'}$ is obtained by adding a $t'$-step (we empirically choose $t' = 1$) noise to the reference video. We run a $1$-step denoising process $\epsilon_{\theta}(\mathbf{z}_{t'}, {t'}, \emptyset)$. $l=\{1, 2, ..., L\}$ is the layer number of temporal transformer modules. $\emptyset$ represents the null-text prompt.
Here, $\mathbf{F}_l \in \mathbb{R}^{C_l \times F \times H_l \times W_l}$, where $C_l$, $H_l$ and $W_l$ represents the channel number, spatial height and width in the output feature $\mathbf{F}_l$ of the $l$-th temporal attention module.} We then extract the feature correlation patterns from $\mathbf{F}_l$. Rather than examining each pair of features in $\mathbf{F}_l$ that is time-consuming,
we instead select a few key points, which can be collected from human-interactive click, centroids of the box trajectories, tracking with detector, or random selection, to calculate their correlation patterns with features in the other frames.
In our implementation, we use one point from the human-interactive click as the starting point and track it through the video, and omit $l$ in the rest of this subsection for clear description since we operate the same operation in each temporal transformer module.

Given one point $p=(y,x)$ in frame $f \in \{1, 2, ..., F\}$, we can extract its feature $\pmb{f}=\mathbf{F}[:, f, y, x] \in \mathbb{R}^{C}$ from one temporal transformer module.
To obtain the correlation pattern of $p$ across frames, we calculate the similarity between $\pmb{f}$ at $p$ and the feature $\pmb{f}_{(i,j,k)} = \mathbf{F}[:, i, j, k] \in \mathbb{R}^{C}$ of each point in other frames, where $j\in \{1, 2, ..., H\}$, $k \in \{1, 2, ..., W\}$.
The calculation is only conducted on the subsequent frames, \ie, $i \in \{f+1, f+2, ..., F\}$.

We then formulate the correlation pattern $\pmb{\mathcal{M}}$ as follows: %
\begin{equation}
\centering
\begin{aligned}
\pmb{\mathcal{M}}&=\{ \pmb{\mathrm{M}}_{f+1}, \pmb{\mathrm{M}}_{f+2}, ..., \pmb{\mathrm{M}}_F \}, \\
\pmb{\mathrm{M}}_{i}(j,k)&=\frac{\mathrm{exp}(\mathrm{sim}(\pmb{f}\cdot \pmb{f}_{(i,j,k)})/\tau )}{\sum_{h=1}^{H}\sum_{w=1}^{W}\mathrm{exp}(\mathrm{sim}(\pmb{f}\cdot \pmb{f}_{(i,h,w)})/\tau )},
\end{aligned}
\label{eq:motion_trajectory_guidance}
\end{equation}
where $\mathrm{sim}(\pmb{u}, \pmb{v})=\pmb{u}^{\mathrm{T}}\pmb{v}/ \left|\left| \pmb{u} \right|\right| \left|\left| \pmb{v} \right|\right|$ denotes the cosine similarity between two vectors $\pmb{u}$ and $\pmb{v}$. $\tau$ is a temperature hyper-parameter. $\pmb{\mathrm{M}}_{i}(j,k)$ represents the distribution of the temporal coherent in the $i$-th frame for the given point $p$. The higher value of $\pmb{\mathrm{M}}_{i}(j,k)$, the higher semantic correspondence between the point $(j,k)$ in the $i$-th frame and the point $p$ in the $f$-th frame. Here, we use the $Softmax$ to ensure $\pmb{\mathcal{M}}$ focuses on strongly matched features.

\setlength{\textfloatsep}{0.15cm}
\begin{algorithm}[t]
	\begin{footnotesize}
		\SetAlgoLined
        \SetKwInOut{Initialization}{Initialization}
        \SetKwInOut{Output}{Output}
        \SetKwInput{Require}{Require}
        \Require{3D U-Net denoiser $\epsilon_{\theta}$; 3D VAE $Encoder$ and $Decoder$; Maximum timestep $T$; Reference video $x_{ref}$; Target text prompt $y$; The number of gradient guidance steps $n$; Guidance scale $\sigma_t$.
        }
        \Initialization{
        \hspace{0.00cm} i) Compute the latent code  $\mathbf{z}_{ref}$=$Encoder(x_{ref})$; \\
        \hspace{-0.6cm} ii) Compute the inversed noisy latent $\mathbf{z}_T$ from $\mathbf{z}_{ref}$; \\
        \hspace{-0.6cm} or use the trajectory-based noise as in \cite{qiu2024freetraj}; \\
        \hspace{-0.6cm} iii) Compute the inversed noisy latent $\mathbf{z}_{t'}$ from $\mathbf{z}_{ref}$; \\
        \hspace{-0.6cm} iv) Conduct $1$-step denoising $\epsilon_{\theta}(\mathbf{z}_{t'}, {t'}, \emptyset)$ to extract \\
        \hspace{-0.6cm} feature maps from temporal transformer modules; \\
        \hspace{-0.6cm} v) Obtain one point $p$ on a specific frame $f$;\\
        \hspace{-0.6cm} vi) Calculate motion correlation pattern $\pmb{\mathcal{M}}$ for $p$ \\
        \hspace{-0.6cm} according to Eq.~\ref{eq:motion_trajectory_guidance};
	}
        \For{$t=T$ \KwTo $1$}
        {
        \eIf{$T - t ~\textless n$}
        {
        Conduct $\epsilon_{\theta}(\mathbf{z}_t, t, y)$ to extract feature maps from temporal transfer modules; \\
        Calculate motion correlation pattern $\pmb{\mathcal{M}}'$ for $p$ according to Eq.~\ref{eq:motion_trajectory_guidance}; \\
        Calculate consistency loss $\mathcal{L}_c$ according to Eq.~\ref{eq:consistency_loss}; \\
        Calculate noise estimation according to Eq.~\ref{eq:guidance} with  $\hat{\epsilon}_\theta({\mathbf z}_t, t, y) := \epsilon_{\theta}(\mathbf{z}_t, t, y) + \sigma_t \nabla_{\mathbf{z}_t} \mathcal{L}_c(\mathbf{z}_t)$;
        }
        {
        $\hat{\epsilon}_\theta({\mathbf z}_t, t, y) := \epsilon_{\theta}(\mathbf{z}_t, t, y)$
        }
        Calculate latent code $\mathbf{z}_{t-1}$;
        }

        $x_0 = Decoder(\mathbf{z}_0)$
        
        \Output{$x_0$}
	\caption{Our training-free motion trajectory guided denoising process.}\label{algo}
	\end{footnotesize}
\end{algorithm}
\setlength{\floatsep}{0.15cm}

\myparagraph{Remark:} 

\noindent i) The motion pattern of a key point can be understood as a soft trajectory. If a point has a high matching score (\eg, cosine similarity) with a point in another frame while its matching scores with other locations remain low, then $\pmb{\mathrm{M}}_i$ approaches a one-hot vector. This simplifies to a standard trajectory, where one point is distinctly matched to another point in a different frame. However, in diffusion models—particularly when using features from large time steps or specific layers—a distinct matching point often does not exist. In such cases, $\pmb{\mathrm{M}}_i$ becomes more uniform, resembling a relaxed trajectory. Empirically, we find it beneficial to include features from various time steps and blocks, even if these features may not exhibit precise semantic correspondence.

\noindent ii) Depending on the video generation model~\cite{video_crafter,zeroscope}, features from different spatial locations at different frames may not directly interact; for example, some models do not implement full spatial-temporal attention. As a result, the motion pattern $\pmb{\mathrm{M}}_i$ differs from the attention maps typically extracted from a diffusion model.

\subsection{Frame-to-Frame Motion Consistency Loss}
\label{sec:consistency_loss}

With the extracted motion pattern $\pmb{\mathcal{M}}$ from the reference video, we further consider how to transfer this motion pattern to
the generated video.%
We design a frame-to-frame consistency loss $\mathcal{L}_c$ to measure how well the motion trajectory $\pmb{\mathcal{M}}'$, derived from the output feature from the noisy latent $\mathbf{z}_t$, is consist with the motion trajectory guidance $\pmb{\mathcal{M}}$.
Note that $\mathbf{z}_t$ is the $t$th-timestep latent in the denoising process, initialized with $\mathbf{z}_T$ from the inversion of the reference video.
For each timestep, we minimize the loss $\mathcal{L}_c$:
\begin{equation}
\centering
\begin{aligned}
\mathcal{L}_c = \sum_{f=1}^F\sum_{i=f+1}^{F}|| \pmb{\mathrm{M}}_{i}^{'} - \pmb{\mathrm{M}}_{i}||_2^{2},
\end{aligned}
\label{eq:consistency_loss}
\end{equation}
where $\pmb{\mathcal{M}}^{'}=\{\pmb{\mathrm{M}}^{'}_{f+1}, \pmb{\mathrm{M}}^{'}_{f+2}, ..., \pmb{\mathrm{M}}^{'}_{F}\}$ shares the same calculation as $\pmb{\mathcal{M}}$, while is extracted at the current time step from the noisy latent.
Here, we simply use the same $\pmb{\mathcal{M}}$ as the motion guidance for all time steps\footnote{It is possible to use different $\pmb{\mathcal{M}}$ in different time steps, which we leave it for the future since it may need more time consumption.}.

Then we can replace $\mathcal{L}_e$ in the noise estimation in Eq.~\ref{eq:guidance} as our proposed $\mathcal{L}_c$, formulated as $\hat{\epsilon}_\theta({\mathbf z}_t, t, y) := \epsilon_{\theta}(\mathbf{z}_t, t, y) + \sigma_t \nabla_{\mathbf{z}_t} \mathcal{L}_c(\mathbf{z}_t).$
The overall details of our method is in Algorithm~\ref{algo}, which is a training-free technique without any parameter update.
Therefore, it is flexible to use our method to any updated or state-of-the-art foundation models.

\myparagraph{Variant on trajectory-controllable video generation.} The method above is based on a given reference video. When meeting trajectory based methods~\cite{wu2023tune,zhang2024tora}, \ie, trajectory-controllable video generation, we follow the the original noise initialization in their methods, such as \cite{qiu2024freetraj}. 
For the motion pattern extraction in Section~\ref{sec:motion_traj_extract}, we first build a box moving video with the given trajectory, which is used as the reference video $x_{ref}$ in Algorithm~\ref{algo}.
Therefore, our method supports both reference video based and trajectory based approaches.

%% file: sec/4_experiment.tex
\section{Experiments}
\label{sec:experiments}

\subsection{Datasets and Evaluation Metrics}
\myparagraph{Datasets.} We conduct the comparison experiments on three benchmarks.
The first one is the open-source benchmark LOVEU-TGVE in the CVPR 2023 competition~\cite{wu2023cvpr}. This dataset comprises 76 videos,
each originally associated with 4 editing text prompts, which will generate 304 videos.
The second one is 56 prompts used in FreeTraj~\cite{qiu2024freetraj}, which are mostly extended from previous baselines~\cite{jain2024peekaboo,trailblazer}.
For each prompt, we initialize 16 random noises with 8 different trajectories and 2 random initial noises, resulting in total 896 generated videos.
We also conduct experiments on UCF Sports Action~\cite{soomro2015action} for motion customization following the adapted text prompts in MotionDirector~\cite{zhao2025motiondirector}.

\begin{table*}[t]
\begin{center}
\setlength{\tabcolsep}{4pt} %
\renewcommand{\arraystretch}{1.1} %
\scalebox{0.85}
{\begin{tabular}{lcccccccc}
\hline
\multicolumn{5}{c}{Automatic Evaluations} &  \multicolumn{4}{|c}{Human Evaluations}  \\ 
\multicolumn{1}{l}{Method}  &\multicolumn{1}{c}{CLIP-SIM ($\uparrow$)} &\multicolumn{1}{c}{CLIP-SIM-GTBox ($\uparrow$)} &\multicolumn{1}{c}{mIoU ($\uparrow$)} &\multicolumn{1}{c}{CD ($\downarrow$)} & \multicolumn{1}{|c}{Comparison} & Trajectory Align & Text Align & Video Quality \\
\hline
Peekaboo~\cite{jain2024peekaboo} & 0.942  & 0.869  & 0.143 & 0.23 & \multicolumn{1}{|c}{ \vs Ours} & 10.94 \vs 89.06 & 24.44 \vs 75.56 & 30.20 \vs 69.80 \\
FreeTraj~\cite{qiu2024freetraj} & \textbf{0.951}  & 0.886 & 0.268 & 0.11  & \multicolumn{1}{|c}{ \vs Ours} & 47.88 \vs 52.12 & 49.11 \vs 50.89 & 42.66 \vs 57.34  \\
Ours & 0.947  &\textbf{0.889}  &\textbf{0.272} & \textbf{0.11} & \multicolumn{1}{|c}{ -} &  - & - & - \\
\hline
\end{tabular}}
\caption{\textbf{Quantitative comparison of trajectory control.} Automatic and human evaluations results with the trajectory based videos.
We re-implement Peekaboo~\cite{jain2024peekaboo} and FreeTraj~\cite{qiu2024freetraj} using their official code with the same prompts as ours. 
Our method achieves competitive performance in metrics about video quality and gains the best scores in metrics that are related to trajectory control. 
}
\vspace{-0.5cm}
\label{tab:traj_compar}
\end{center}
\end{table*}

\begin{figure*}[t]
\centering
\includegraphics[trim =0mm 0mm 0mm 0mm, clip, width=1.0\linewidth]{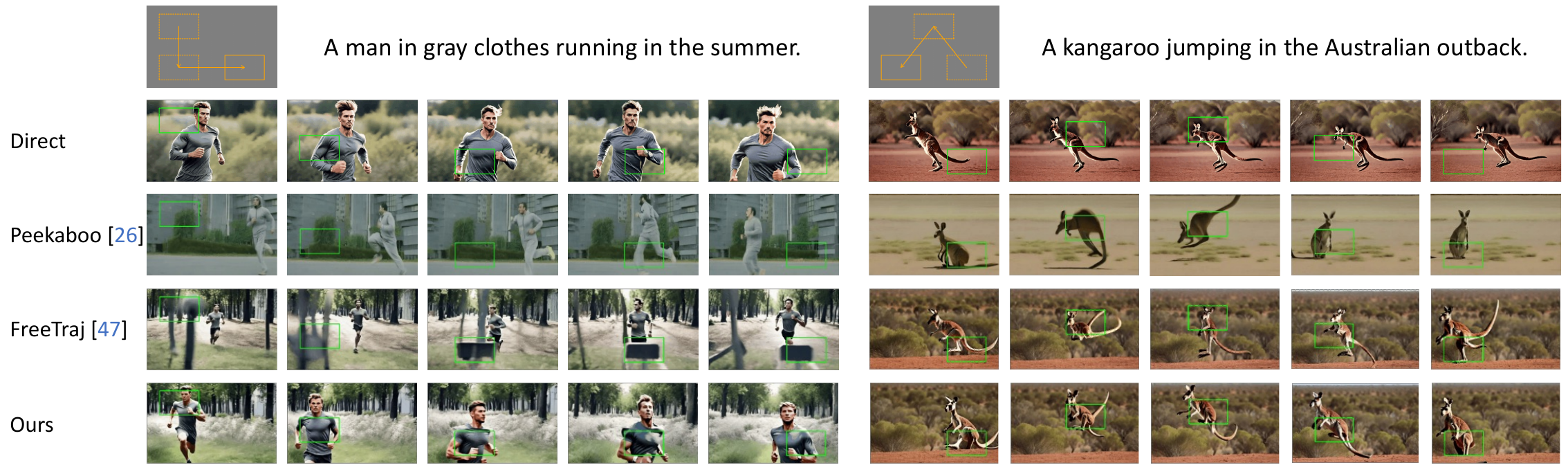}
\caption{\textbf{Qualitative comparison of trajectory control.}
We evaluate our method and other trajectory based approaches, \ie, Peekaboo~\cite{jain2024peekaboo} and FreeTraj~\cite{qiu2024freetraj}.
The ``Direct'' means the direct inference with random noise and no other guidance.
We use the same initial noises as in \cite{qiu2024freetraj} for better visual comparison.
Our method shows better ability on trajectory follow and temporal coherent consistency.
}
\vspace{-0.3cm}
\label{fig:traj_compar}
\end{figure*}

\myparagraph{Evaluation metrics.} We compare the proposed method with reference video based MotionDirector~\cite{zhao2025motiondirector} that is a training-based controllable generation method.
Following the LOVEU-TGVE competition~\cite{wu2023cvpr}, we use the average CLIP score~\cite{hessel2021clipscore} between the diverse text prompts and all frames of the generated videos to compute the appearance diversity (AP), and the average CLIP score between frames to compute the temporal consistency (TC), and the average PickScore~\cite{kirstain2023pick} between all frames of output videos to compute the Pick Score (PS).

We also compare a trajectory-controllable method FreeTraj~\cite{qiu2024freetraj}. 
We report CLIP Similarity (CLIP-SIM)~\cite{radford2021learning} to measure the semantic similarity among frames, and propose to use CLIP Similarity in the ground-truth trajectory boxes (CLIP-SIM-GTBox) to measure the semantic similarity in the trajectory boxes among frames, evaluating the trajectory following and temporal coherent ability.
Following \cite{qiu2024freetraj}, we also alculate the Mean Intersection of Union (mIoU) and Centroid Distance (CD) to evaluate the trajectory alignment by obtaining the bounding box of the synthesized objects with OWL-ViT-large~\cite{minderer2022simple} detector.

\myparagraph{Human Preference.} We conduct user study by shuffling our generated videos and videos from MotionDirector~\cite{zhao2025motiondirector} or FreeTraj~\cite{qiu2024freetraj}.
A total of 25 users were asked to pick the best one video in three dimensions, \ie, video-text alignment (text align), the trajectory or motion alignment (trajectory / motion align), and video quality, respectively.
To simplify the comparison for raters, users are asked to compare the results pairwise and select their preferred one.
The pairwise numbers ``$x_1$ \vs $x_2$'' in the human evaluation represents that $x_1\%$ proportion that generated videos from the compared method are preferred, and $x_2\%$ proportion that generated videos from ours are preferred.

\subsection{Implementation Details}
For fair comparison, we use VideoCrafter~\cite{video_crafter} and ZeroScope~\cite{zeroscope} as the pre-trained video model following \cite{qiu2024freetraj} and \cite{zhao2025motiondirector} for the trajectory based setting and the reference video based setting, respectively.
By default, we set the weighting schedule $\sigma_t$ to 10000.0, and temperature in Eq.~\ref{eq:motion_trajectory_guidance} to 10.0.
The number of gradient guidance steps $n=T$, where $T=50$ for trajectory-based methods and $T=30$ for reference video based methods.
To generate the sparse points, we design a simple GUI to collect points from human-interactive click.
We simply use the centriod point of the box trajectory as the sparse point.
More details are in the supplementary material.

\subsection{Evaluation of Trajectory Control}
We compare our method to other training-free trajectory-controllable video generation method with diffusion models, including Peekaboo~\citep{jain2024peekaboo} and FreeTraj~\citep{qiu2024freetraj}. 
Our method uses the same noise initialization as in FreeTraj~\citep{qiu2024freetraj} for fair comparison.

\myparagraph{Quantitative evaluation.}
The left part in Table~\ref{tab:traj_compar} shows the quantitative evaluations on these training-free trajectory-controllable video generation methods.
Our method and FreeTraj~\cite{qiu2024freetraj} uses trajectory-based noise initialization and attention operation, improves the model's ability of trajectory following.
With the same initialization as in \cite{qiu2024freetraj}, our method with the motion pattern consistency guidance can further improves the control ability of trajectory, \eg, improves 0.4\% on mIoU and 0.3\% on CLIP-SIM-GTBox, showing the strong ability of our method on improving temporal coherent.
\xinyu{Meanwhile, we find the CLIP-SIM, which are the references for video quality, are slightly worse than FreeTraj.
The underlying reason is that FreeTraj sometimes generate videos with slight motion, failing to follow trajectories, \eg, left part in Figure~\ref{fig:traj_compar}; so that the the similarity among frames will be improved.
The similar observation is also in FreeTraj that the naive baseline has the best CLIP-SIM since the model tends to generate static videos. 
We then crop the videos in the ground-truth trajectory boxes to show whether the temporal content in the GT box is consistent or not, \ie, calculating CLIP-SIM-GTBox.
It clearly shows that our method can achieve the best performance since our method can maintain the temporal coherent and follow the trajectory movement.
}

\begin{figure*}[t]
\centering
\includegraphics[trim =0mm 0mm 0mm 0mm, clip, width=1.0\linewidth]{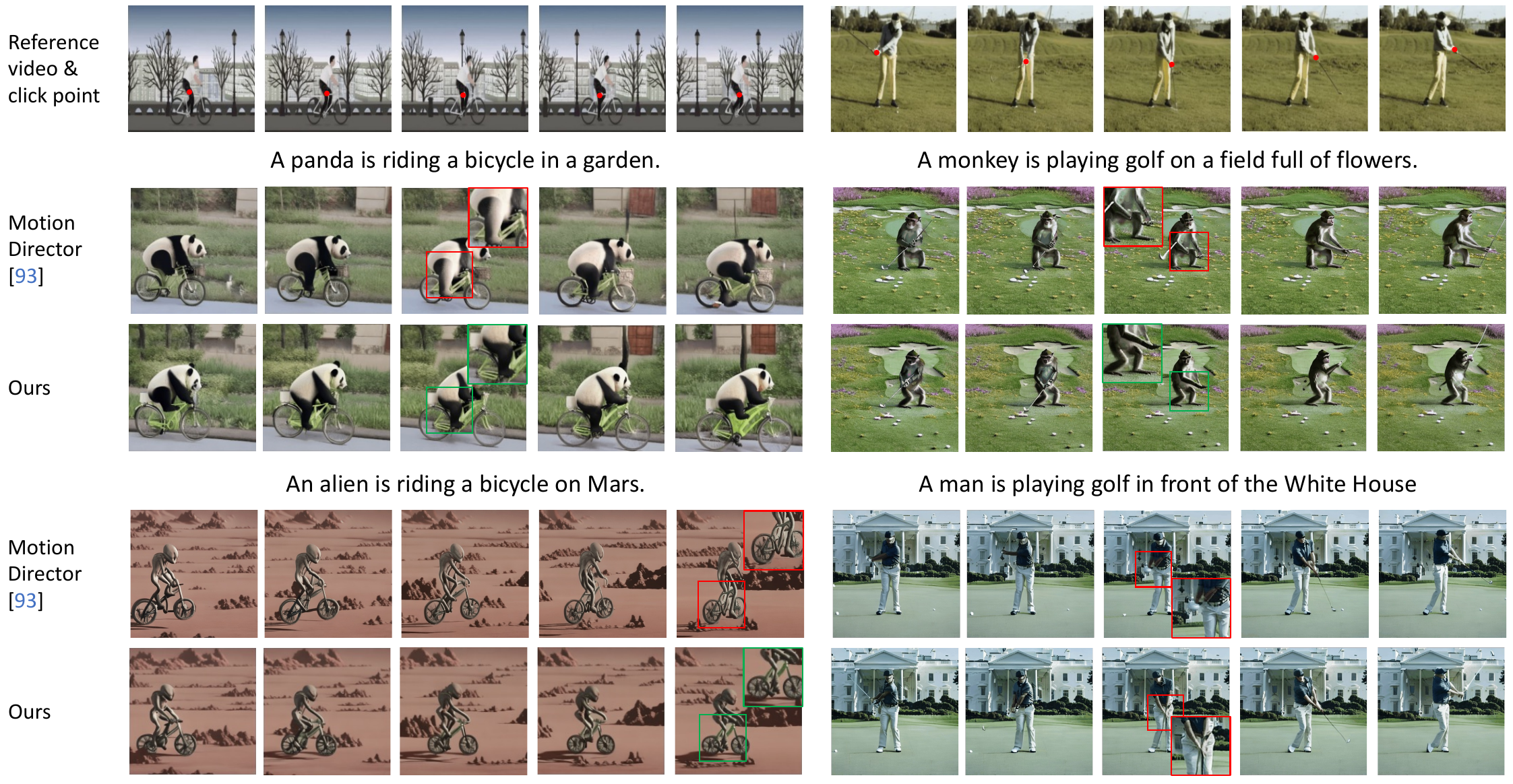}
\vspace{-0.5cm}
\caption{\textbf{Qualitative comparison of reference video control.} We evaluate our method and MotionDirector~\cite{zhao2025motiondirector}. The \textcolor{red}{red} circle represents the given point clicked by users. The \textcolor{red}{red} and \textcolor{darkgreen}{green} rectangle are highlight areas to show the temporal coherent clearly. We keep the initial noises same in \cite{qiu2024freetraj} and our method for fair comparison.}
\label{fig:refer_video_compar}
\end{figure*}

\begin{table*}[t]
\centering
\setlength{\tabcolsep}{6pt} %
\renewcommand{\arraystretch}{1.1} %
\scalebox{0.85}{
\begin{tabular}{lccclccc}
\hline
\multicolumn{4}{c}{Automatic Evaluations} &  \multicolumn{4}{|c}{Human Evaluations}  \\ 
\hline
\multicolumn{1}{l}{\multirow{1}{*}{Method}} & AD ($\uparrow$) & TC ($\uparrow$) & PS ($\uparrow$) & \multicolumn{1}{|c}{\multirow{1}{*}{Comparison}} & Text Align & Motion Align & Video Quality \\ 
\hline
MotionDirector~\cite{zhao2025motiondirector} & 25.95  & 93.1 & 20.37 & \multicolumn{1}{|c}{\vs Ours} & 48.68 \vs 51.32 & 46.05 \vs 53.95 & 44.74 \vs 55.26 \\
Ours & 25.74 & 93.3 & 20.34 & \multicolumn{1}{|c}{-} & - & - & - \\
\hline
\end{tabular}
}
\caption{\textbf{Quantitative comparison of reference video control.} Automatic and human evaluations results of motion customization on single videos. 
We re-implement MotionDirector~\cite{zhao2025motiondirector} using the official code with the same prompts as ours. 
}
\vspace{-0.4cm}
\label{tab:refer_video_compar}
\end{table*}

\myparagraph{Quantitative human evaluation.}
We also conduct a user study to evaluate our results based on human subjective preference. 
Users are instructed to select the preferred generated videos from two candidates in three evaluation metrics: trajectory alignment, video-text alignment, and video quality.
The results are shown in the right part in Table~\ref{tab:traj_compar}.
It clearly shows that our approach outperforms all other trajectory-based methods by a significant margin in all metrics.
Notably, our method can further improve the trajectory alignment even applying the motion pattern consistency guidance on the relative strong method FreeTraj, and improve a lot on the video quality due to the better temporal coherent consistency.

\myparagraph{Qualitative evaluation.}
As shown in Figure~\ref{fig:traj_compar}, Peekaboo~\cite{jain2024peekaboo} has the worst control to follow the given trajectory.
FreeTraj~\cite{qiu2024freetraj} can follow the given trajectory but not always precisely.
In addition, generated videos from FreeTraj sometimes have weird artifacts and inconsistent content among frames, \eg, there is a box artifact in the left example, and the kangaroo has two tails in the last frame.
Our method can not only succeed in driving the target object following the given trajectories, but also maintain consistent temporal coherent.

\subsection{Evaluation of Reference Video Control}
We compare our method to the training-based reference video based video generation method, MotionDirector~\cite{zhao2025motiondirector}. 
We use the same noise initialization as in MotionDirector~\citep{qiu2024freetraj} for fair comparison, \ie, inversion noise initialization as described in Section~\ref{sec:initial_noise}.

\myparagraph{Quantitative evaluation.}
In Table.~\ref{tab:refer_video_compar}, we refer to the alignment between the generated videos and the 4 editing text prompts as the appearance diversity. 
The results show that our method achieves competitive performance compared with the reference video based method, MotionDirector~\cite{zhao2025motiondirector}, which requires additional training.
Our method and MotionDirector achieve competitive performance on appearance diversity (AD) and Pick Score (PS), while the latter has slightly better result.
The underlying reason is that the our method make all 4 generated video to be similar to the motion of reference videos, resulting in the reduction of diversity.
In contrast, with motion pattern consistency guidance, our method achieves better result than MotionDirector on temporal consistency (TC) metric.
To intuitively reflect the effectiveness of our method,  
we add the user study to evaluate the quality of generated videos by humans.

\myparagraph{Quantitative human evaluation.}
We evaluate our results based on human subjective preference, with the selection of the preferred generated videos from two candidates in three evaluation metrics: video-text alignment, motion alignment with the reference video, and video quality.
The results are shown in the right part in Table~\ref{tab:refer_video_compar}.
Since generated videos from MotionDirector sometimes have appearance artifacts, or inconsistent motion among frames, the results are lower than that of our method on all three human evaluation metrics.
It shows that our method achieves high temporal coherent among frames.

\myparagraph{Qualitative evaluation.}
We take a set of videos from UCF Sport Action~\cite{soomro2015action} as the reference videos for the qualitative comparison. To compare MotionDirector~\cite{zhao2025motiondirector} and proposed method fairly, we feed the same initial noise\footnote{We find that MotionDirector~\cite{zhao2025motiondirector} is sensitive to the random seed, \ie, the initial noise. To show the effectiveness of our method, we here use the same random seed for fair comparison.}, consisting of the weighted sum of the inversion noise from the reference video and a random Gaussian noise, to both \cite{zhao2025motiondirector} and our method to generate videos.

\xinyu{As shown in Fig.~\ref{fig:refer_video_compar}, the Baseline can correctly generates the appearance but lacks the realistic motion from the reference video.
MotionDirector~\cite{zhao2025motiondirector} can generate the desired motion since temporal loras are trained on the reference video to inject the model.
However, it still suffer the issue of frame-to-frame temporal coherent inconsistency with obvious appearance artifacts.
Our method can draw a similar motion from the reference video while has good consistency among frames, showing the the ability of following the motion from the reference video and temporal consistency.
}

\subsection{Ablation Studies}
\label{sec:ablation}

\begin{figure}[t]
\centering
\includegraphics[trim =0mm 0mm 0mm 0mm, clip, width=1.0\linewidth]{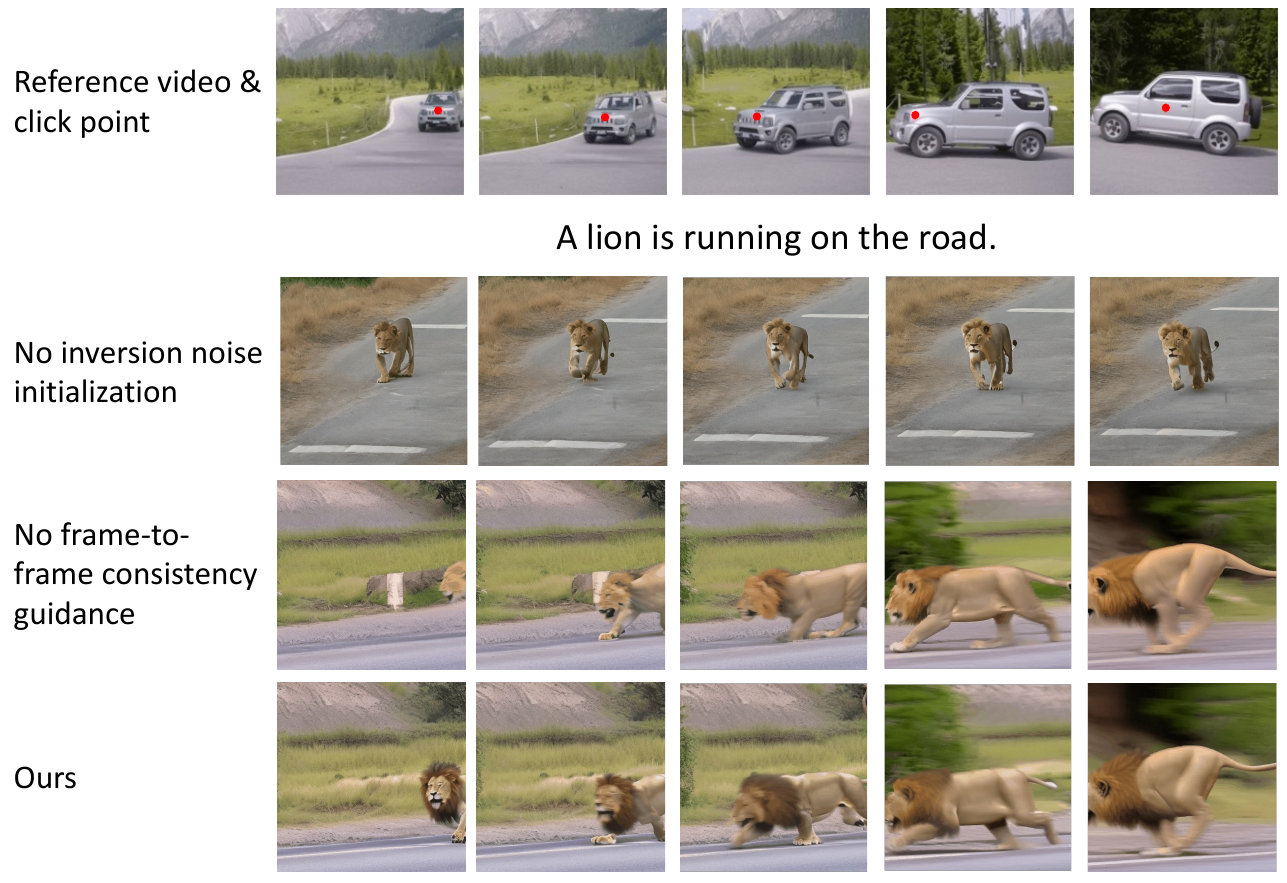}
\caption{\textbf{Ablation study} on each component in our method, including the inversion noise initialization and frame-to-frame consistency guidance.
}
\vspace{0.2cm}
\label{fig:ablation_compar}
\end{figure}

\myparagraph{Impact on inversion noise initialization.}
As shown in Figure~\ref{fig:ablation_compar}, when not using the inversion noise initialization, our method reduces to apply the frame-to-frame consistency guidance on a random noise.
It can follow a rough route of the reference car motion, while failing to simulate the large motion with appearance variance, \ie, car turn.
It shows that the inversion noise initialization could provide the motion trajectory implicitly.

\myparagraph{Impact on frame-to-frame consistency guidance with $\mathcal{L}_c$.}~
If there is no frame-to-frame consistency guidance with $\mathcal{L}_c$, our method reduces to \cite{zhao2025motiondirector} as we use its pipeline.
\cite{zhao2025motiondirector} trains temporal LoRAs to learn the motion from the reference video, thus the model can successfully capture the motion movement.
However, the temporal consistency is not good, especially on the content details.
For example, the hairs on the lion's head vary among frames.
With our proposed frame-to-frame consistency guidance, the temporal coherent clearly improves, showing the effectiveness of our method.
\textcolor{black}{An interesting observation from Figure \ref{fig:ablation_compar} is that the tracking error in the last frame, where the point is tracked on the car body instead of the car front, does not affect temporal coherence. This is because our approach characterizes motion through feature correlations between key points and local features across multiple frames, allowing minor errors in individual frames to be tolerated.}

\begin{figure}[t]
\centering
\includegraphics[trim =0mm 0mm 0mm 0mm, clip, width=1.0\linewidth]{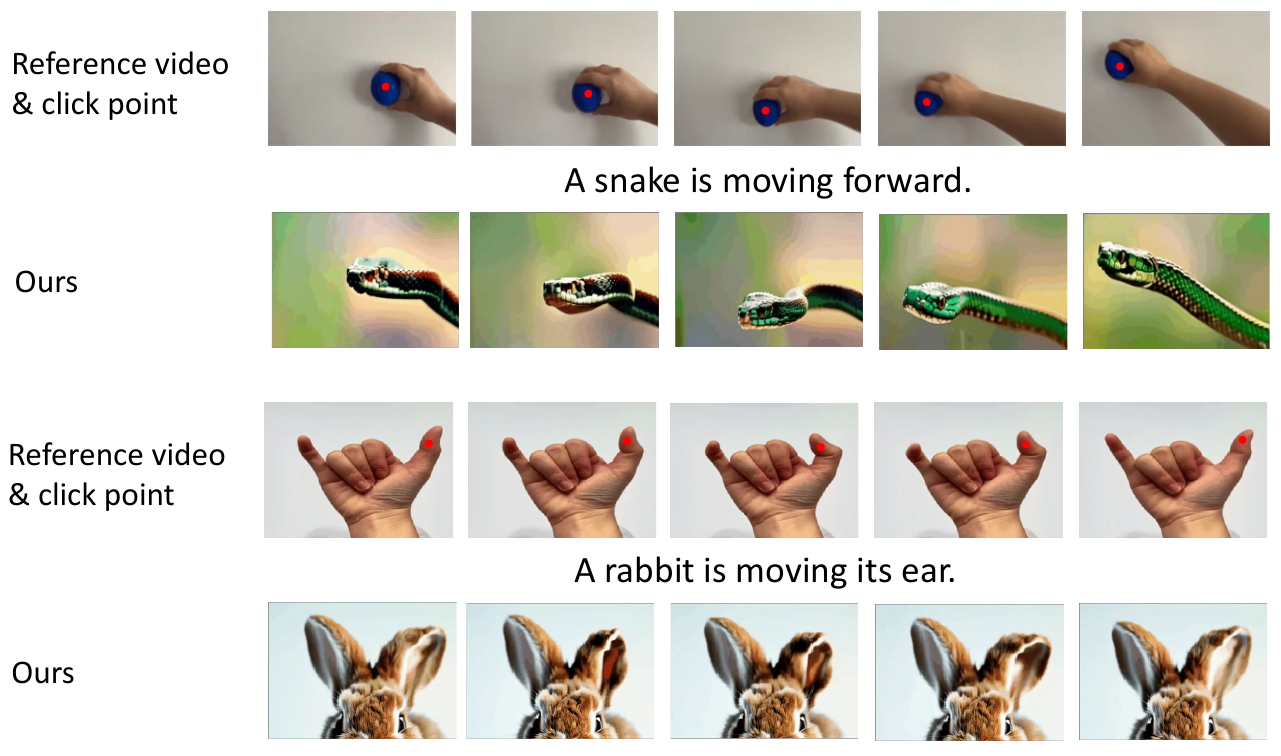}
\caption{\textbf{Gesture simulation.} We use two gestures captured from the camera to simulate animal's movement. Our method can successfully generate the video with accurate ear moving of rabbit and body moving of snake when the given point is on the finger and ball. (Best view in video in the project link.)
}
\vspace{0.2cm}
\label{fig:gesture_simulate}
\end{figure}

\myparagraph{Gesture simulation.}
We also conduct an interesting experiment to verify the effectiveness of our method on gesture simulation. 
We use iPhone 15 pro max to take two videos of hand moving; one is with two fingers bent, and another one with holding a ball moving.
We find that our method can successfully simulate the motion of fingers to that of the rabbit's ear and snake's head.
It will be useful when the target video is hard to be captured, \eg, when a child talks to a snake in the movie (a similar scene as in Harry Hotter), we can use gesture to simulate the snake's behavior and generate it.

%% file: sec/5_conclusion.tex
\section{Conclusion}
\label{sec:conclusion}
This paper addresses the challenge of achieving temporal consistency in motion-guided video generation. We propose a simple yet effective training-free approach that combines inversion noise initialization with a novel motion consistency loss to improve both temporal coherence and motion accuracy. By leveraging inter-frame motion feature pattern correlations at sparse points, our method transfers motion from reference videos or trajectories by replicating these patterns in the generated videos. Extensive qualitative and quantitative experiments demonstrate the effectiveness of our approach in enhancing temporal consistency.

%% file: sec/X_suppl.tex
\clearpage
\setcounter{page}{1}
\maketitlesupplementary

\section{Implementation Details}
\label{sec:more_details}
\myparagraph{Hyperparameters.}
For trajectory control, we use VideoCrafter~\cite{video_crafter} as the pre-trained text-to-video model and generate images at a resolution of $320 \times 512$, following \cite{qiu2024freetraj}. Based on the box trajectory, we simply use the centroid point of the box as the sparse point. Additionally, the box trajectory with a black box and white background is used as the reference video to extract the motion correlation pattern $\pmb{\mathcal{M}}$ in Eq.~\ref{eq:motion_trajectory_guidance}. The initial noise and random seed are kept same in \cite{qiu2024freetraj} and ours for a fair comparison.

For reference video control, we use ZeroScope~\cite{zeroscope} as the pre-trained video model and generate images at a resolution of $384 \times 384$, following \cite{zhao2025motiondirector}. The initial noise is obtained by inverting the reference video (refer to Section~\ref{sec:initial_noise}) and adding sum-weighted random noise, as the implementation in \cite{zhao2025motiondirector}. The DDIM~\cite{ddim} inversion step is set to $30$ to compute $\mathbf{z}_{T}$. Sparse points in the first frame are extracted through human-interactive clicks and tracked throughout the video for subsequent frames. For fair comparison, we use the same random seed in \cite{zhao2025motiondirector} and our method.

By default, we set $\sigma_t = 10000.0$ and $\tau = 10.0$ in Eq.~\ref{eq:motion_trajectory_guidance}. The frame count $F$ is set to $16$ for trajectory control-based methods and $32$ for reference video-based methods. To conserve VRAM, we employ mixed precision inference using FP16. All experiments are conducted on an NVIDIA L40 graphics card with 48GB of GPU memory\footnote{In theory, a GPU with 32GB of memory is sufficient for all experiments in this paper}. The classifier-free guidance scale is set to $12$, and the number of gradient guidance steps for motion consistency is $n=50$ for trajectory-based methods and $n=30$ for reference video-based methods.

\begin{figure}[t]
\centering
\includegraphics[trim =0mm 0mm 0mm 0mm, clip, width=1.0\linewidth]{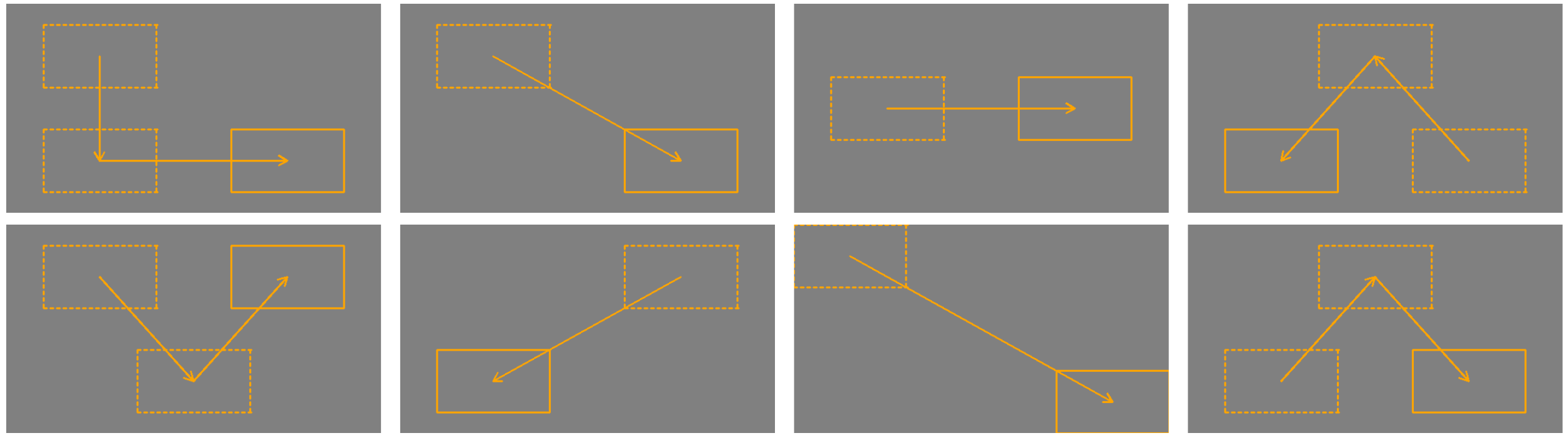}
\vspace{-0.5cm}
\caption{\textbf{The $8$ different box trajectories} in the evaluation of trajectory control methods. 
}
\label{fig:box_traj}
\end{figure}

\myparagraph{Quantitative comparison.} 
For trajectory control, we utilize $56$ prompts following \cite{qiu2024freetraj}, where each prompt is applied to $8$ different trajectories with $2$ random initial noises. This results in a total of $896$ generated videos. All prompts are identical to those provided in the supplementary material of \cite{qiu2024freetraj}. However, the trajectories and random initial noises were re-implemented by us, as they were not publicly available in the original project. The $8$ trajectories are shown in Figure~\ref{fig:box_traj}.

For reference video control, we adopt the open-sourced benchmark released by the LOVEU-TGVE competition at CVPR 2023~\cite{wu2023cvpr}, following the motion customization setting on a single video as described in \cite{zhao2025motiondirector}. The dataset includes $76$ videos, with each video paired with $4$ editing text prompts focusing on object, background, style, and multiple changes. This results in a total of $304$ generated videos. The additional $3$ prompts with large appearance changes for each video in \cite{zhao2025motiondirector} are not available; therefore, we re-implement \cite{zhao2025motiondirector} on the original dataset.

\myparagraph{Details of human preference.}
We use the Gradio\footnote{Abid, Abubakar, Ali Abdalla, Ali Abid, Dawood Khan, Abdulrahman Alfozan, and James Zou. ``Gradio: Hassle-free sharing and testing of ML models in the wild.'' arXiv preprint arXiv:1906.02569 (2019).} toolbox to build a web interface that allows annotators to evaluate the generated videos across three factors:

\begin{enumerate}[i)]
    \item \textbf{Video-text alignment}: Which video better matches the caption ``text prompt''?
    \item \textbf{Trajectory or motion alignment}: Which video more closely follows or aligns with the trajectory or motion of the reference video? (Focus solely on motion information, disregarding appearance or style quality.)
    \item \textbf{Video quality}: Which video is smoother, exhibits less flicker, and is more free from artifacts?
\end{enumerate}

\begin{figure*}[t]
\centering
\includegraphics[trim =0mm 0mm 0mm 0mm, clip, width=1.0\linewidth]{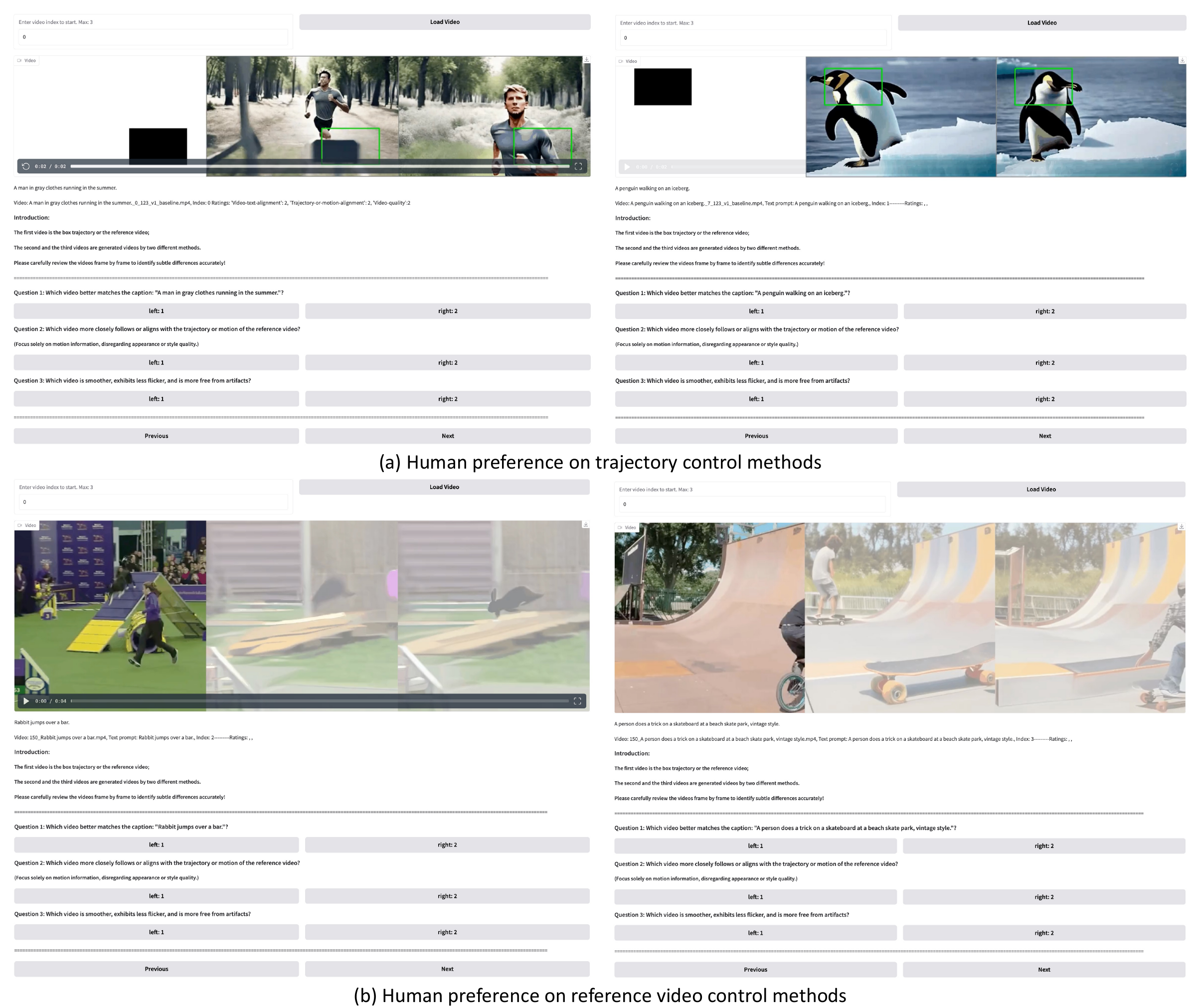}
\vspace{-0.5cm}
\caption{\textbf{Visualization of the website for human preference.} Users are asked to compare the results pairwise and select their preferred one.
}
\vspace{-0.3cm}
\label{fig:human_prefer_ui}
\end{figure*}

An example of the web interface is shown in Figure~\ref{fig:human_prefer_ui}.
We recruited 25 users with undergraduate or postgraduate degrees to participate in the human preference evaluation. For simplicity, users were asked to perform pairwise comparisons and select their preferred results between our method and the compared method with random orders. 
To ensure a fair comparison, we used identical random seeds across our method and the compared method, which may result in differences in local regions and details in the generated videos. Users were instructed to carefully review the videos frame by frame to identify subtle differences accurately.

For trajectory control comparisons, all $896$ generated videos were used. For reference video control methods, $304$ generated videos were included. Among these, $292$ videos were randomly selected from the LOVEU-TGVE~\cite{wu2023cvpr} benchmark, with one prompt randomly chosen from the $4$ prompts available for each video. The remaining $12$ videos were selected from the UCF Sport Action~\cite{soomro2015action} benchmark. The statistical results are presented in Table~\ref{tab:traj_compar} and Table~\ref{tab:refer_video_compar} in the main paper.

\section{More Ablation Studies}

\begin{figure}[t]
\centering
\includegraphics[trim =0mm 0mm 0mm 0mm, clip, width=1.0\linewidth]{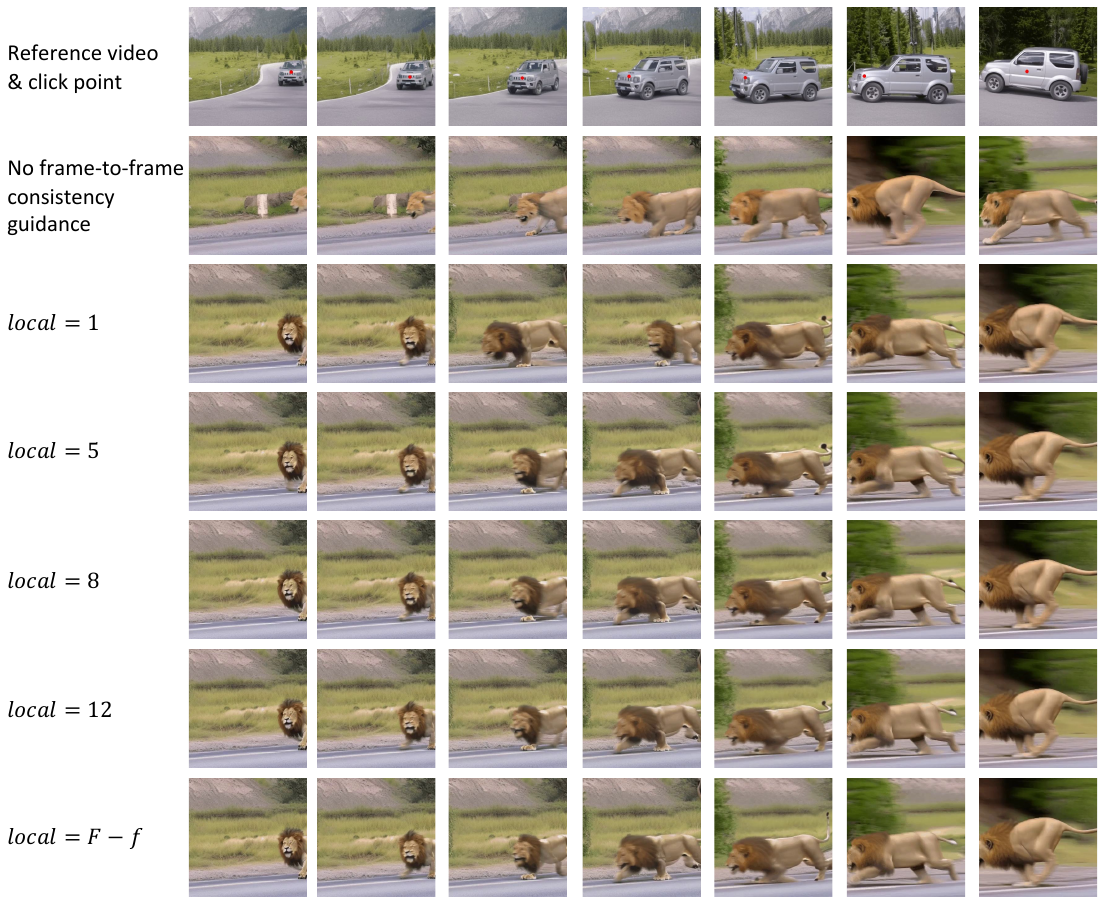}
\vspace{-0.5cm}
\caption{\textbf{The $local$ range} for the calculation of the motion correlation pattern. The text prompt is ``A lion is running on the road.''.
}
\label{fig:local_range_supp}
\end{figure}

\myparagraph{Frame number in the motion pattern extraction.}
In the main paper, we use $F=16$ for trajectory control and $F=32$ for reference video control to calculate the motion correlation pattern $\pmb{\mathcal{M}}$ in Section~\ref{sec:motion_traj_extract} (Line 261, ``The calculation is only conducted on the subsequent frames, ...''). 
Given a point $p=(y,x)$ in frame $f \in \{1, 2, \ldots, F\}$, we adjust the calculation of $\pmb{\mathcal{M}}$ to a smaller range of subsequent frames instead of all subsequent frames. Specifically, the calculation is conducted on subsequent frames, \ie, $i \in \{f+1, f+2, \ldots, f + {local}\}$, where $f + {local} \leq F$. Here, we set ${local} \in \{1, 5, 8, 12, F-f\}$. 
If the number of subsequent frames is less than ${local}$, we use the maximum of ${local}$ and the actual number of subsequent frames instead. Figure~\ref{fig:local_range_supp} shows that as ${local}$ increases to $8$, the motion becomes more similar to the reference video. When ${local}$ increases further, the motion remains stable, while the appearance details improve slightly, such as the lion's tail and claws. By default, we set ${local} = F-f$.

\begin{figure}[t]
\centering
\includegraphics[trim =0mm 0mm 0mm 0mm, clip, width=1.0\linewidth]{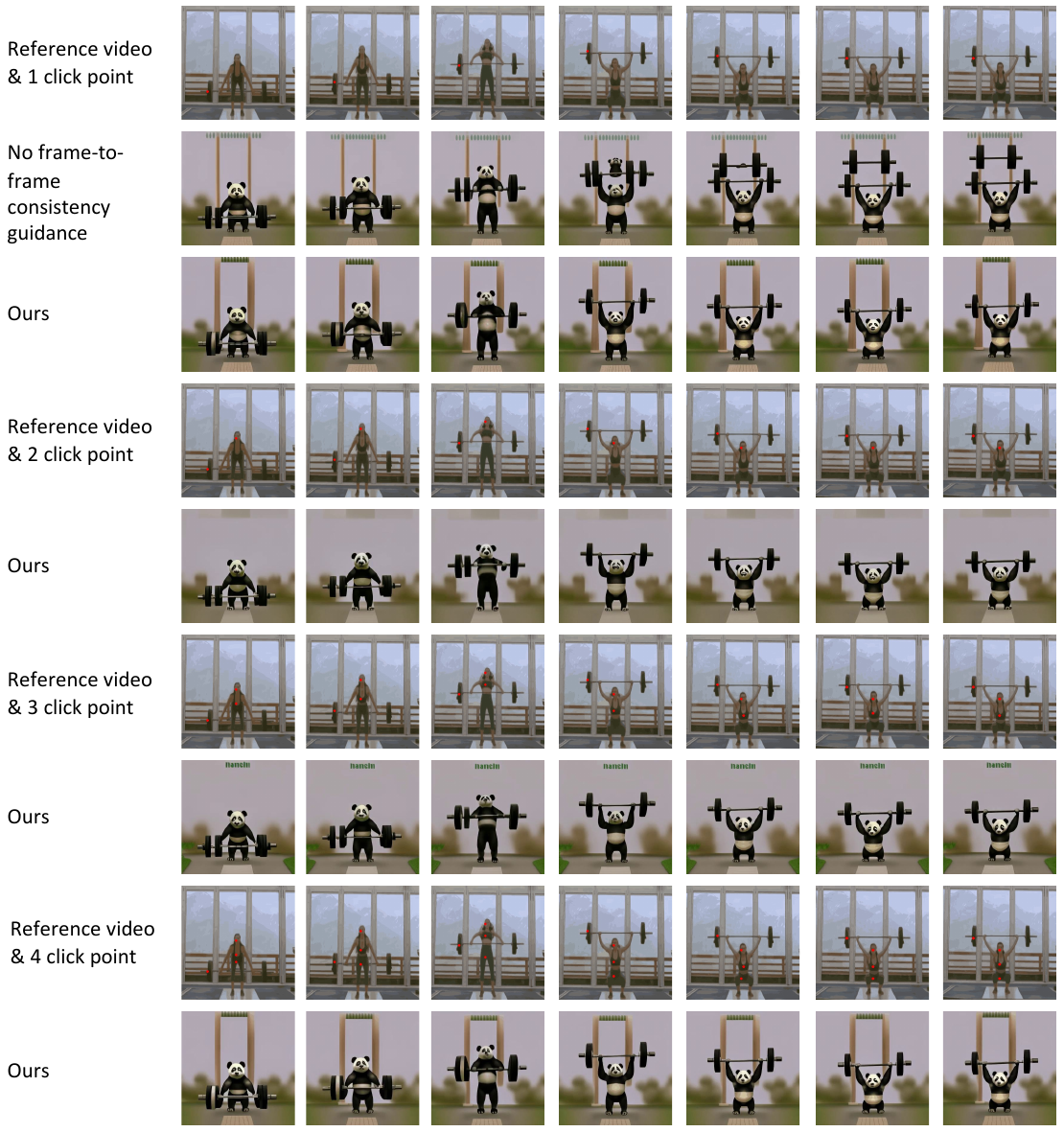}
\vspace{-0.6cm}
\caption{\textbf{The number of sparse points} selected for the calculation of the motion correlation consistency. The text prompt is ``A panda is lifting weights''.
}
\label{fig:selection_point_supp}
\end{figure}

\myparagraph{Sparse point selection.}
In the main paper, we select a single point for motion guidance: the centroid of the box trajectory for trajectory control methods and a human-interactive click for reference video control methods. 
Here, we study the influence of increasing the number of selected points. Figure~\ref{fig:selection_point_supp} demonstrates that with our proposed frame-to-frame consistency guidance, performance consistently improves regardless of the number of sparse points. However, increasing the number of sparse points does not lead to further performance improvements. 
The underlying reason is that our motion patterns are soft correlations between the selected point and other points (refer to Eq.~\ref{eq:motion_trajectory_guidance}). Adding more points may introduce some uncertainty or inaccurate correlations during inference with motion guidance. 
We leave this issue for future research. In this paper, we use a single point by default.

\begin{figure}[t!]
\centering
\includegraphics[trim =0mm 0mm 0mm 0mm, clip, width=1.0\linewidth]{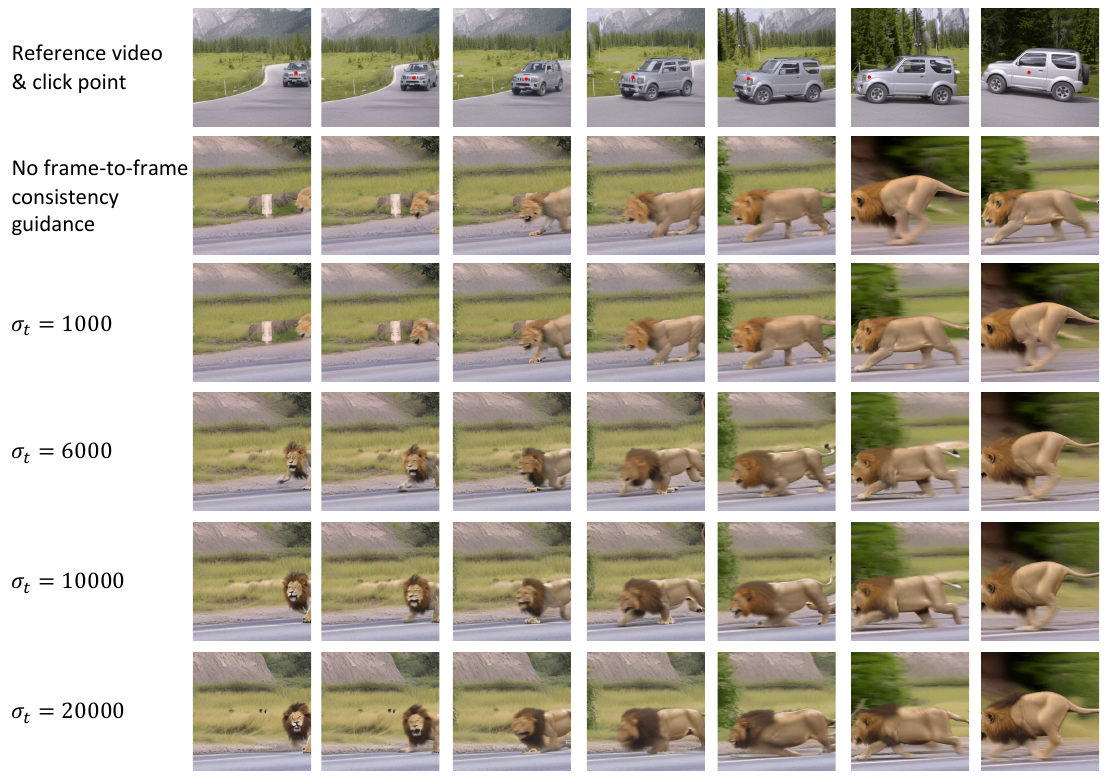}
\vspace{-0.6cm}
\caption{\textbf{The impact on the weight schedule $\sigma_t$} when using the frame-to-frame motion consistency guidance.
}
\label{fig:loss_weight_supp}
\end{figure}

\myparagraph{Impact on the weight schedule $\sigma_t$.}
In the main paper, we set the weighting schedule $\sigma_t = 10000.0$ in Eq.~\ref{eq:guidance} when applying the frame-to-frame motion consistency guidance. 
Here, we ablate to study the impact of the weighting schedule $\sigma_t$. Figure~\ref{fig:loss_weight_supp} illustrates that as $\sigma_t$ increases, the motion becomes more consistent with the reference video. Setting $\sigma_t$ too small reduces its influence on aligning the motion with the reference video, while setting $\sigma_t$ too large will have negative influence on the appearance details, such as the color of the lion's hair. 
When $\sigma_t = 10000.0$, the performance achieves a relatively optimal balance.

\begin{figure}[t!]
\centering
\includegraphics[trim =0mm 0mm 0mm 0mm, clip, width=1.0\linewidth]{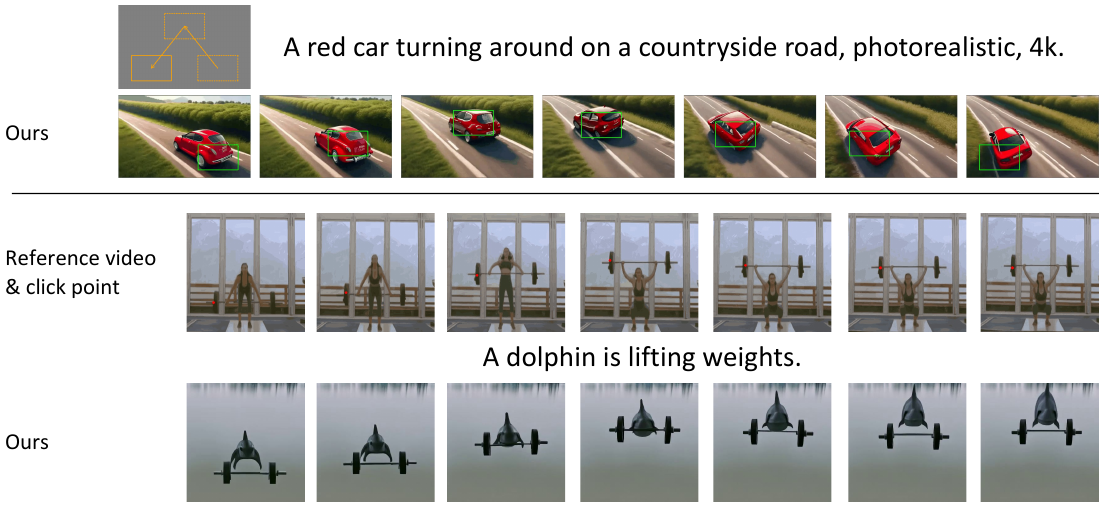}
\vspace{-0.6cm}
\caption{\textbf{Failure case}. The results are suboptimal when the target prompt deviates significantly from the reference video.
}
\label{fig:failure_case_supp}
\end{figure}

\section{Additional Results}

\begin{figure*}[t]
\centering
\includegraphics[trim =0mm 0mm 0mm 0mm, clip, width=0.92\linewidth]{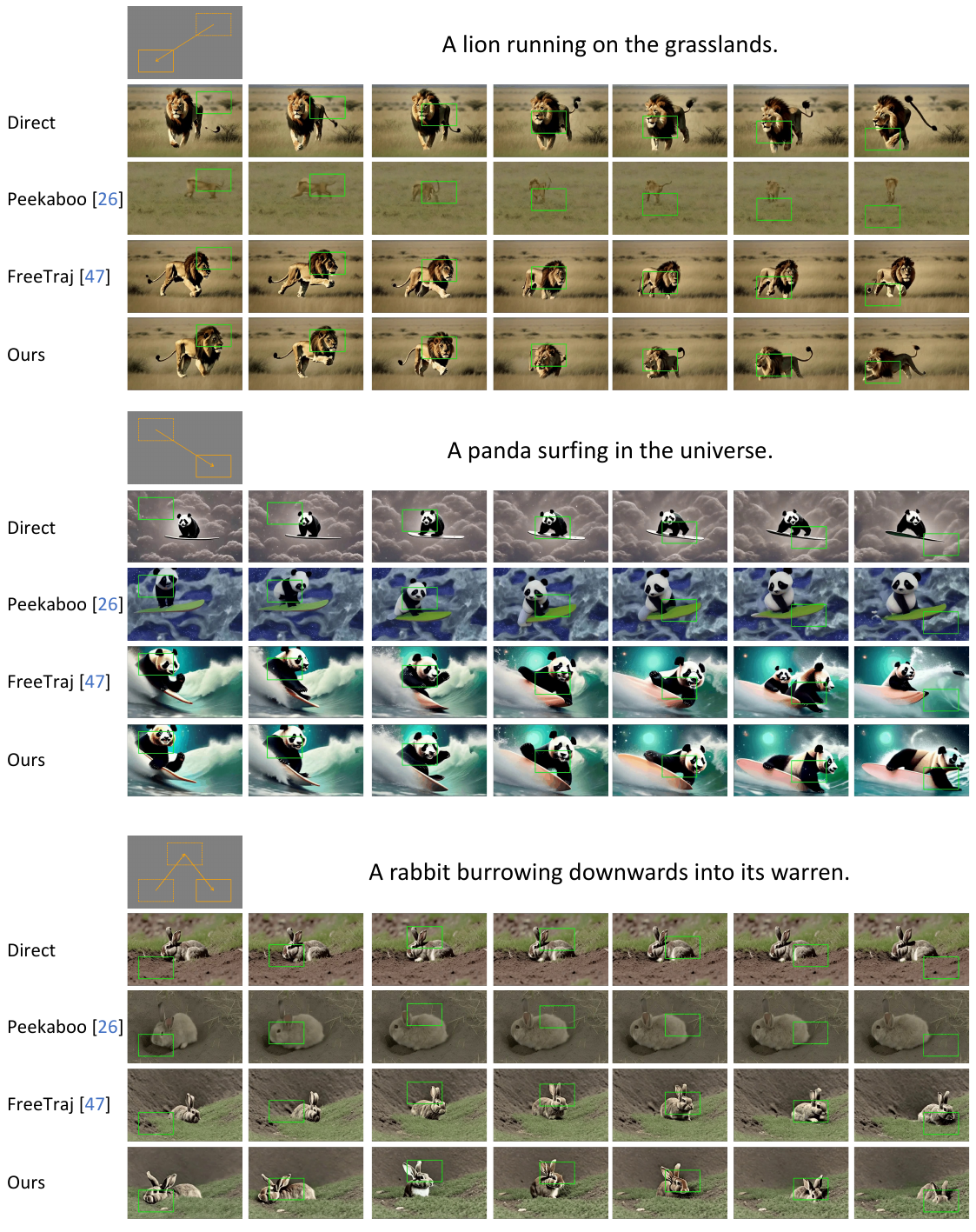}
\caption{\textbf{Qualitative comparison of trajectory control.}
We compare our method with other trajectory based approaches, \ie, Direct inference, Peekaboo~\cite{jain2024peekaboo} and FreeTraj~\cite{qiu2024freetraj}.
The ``Direct'' means the direct inference with random noise and no other guidance.
We use the same initial noises as in \cite{qiu2024freetraj} for better visual comparison.
Our method shows better ability on trajectory follow and temporal coherent consistency. (Best viewed with zoom for finer details.)
}
\label{fig:traj_compar_supp}
\end{figure*}

\begin{figure*}[t]
\centering
\includegraphics[trim =0mm 0mm 0mm 0mm, clip, width=0.9\linewidth]{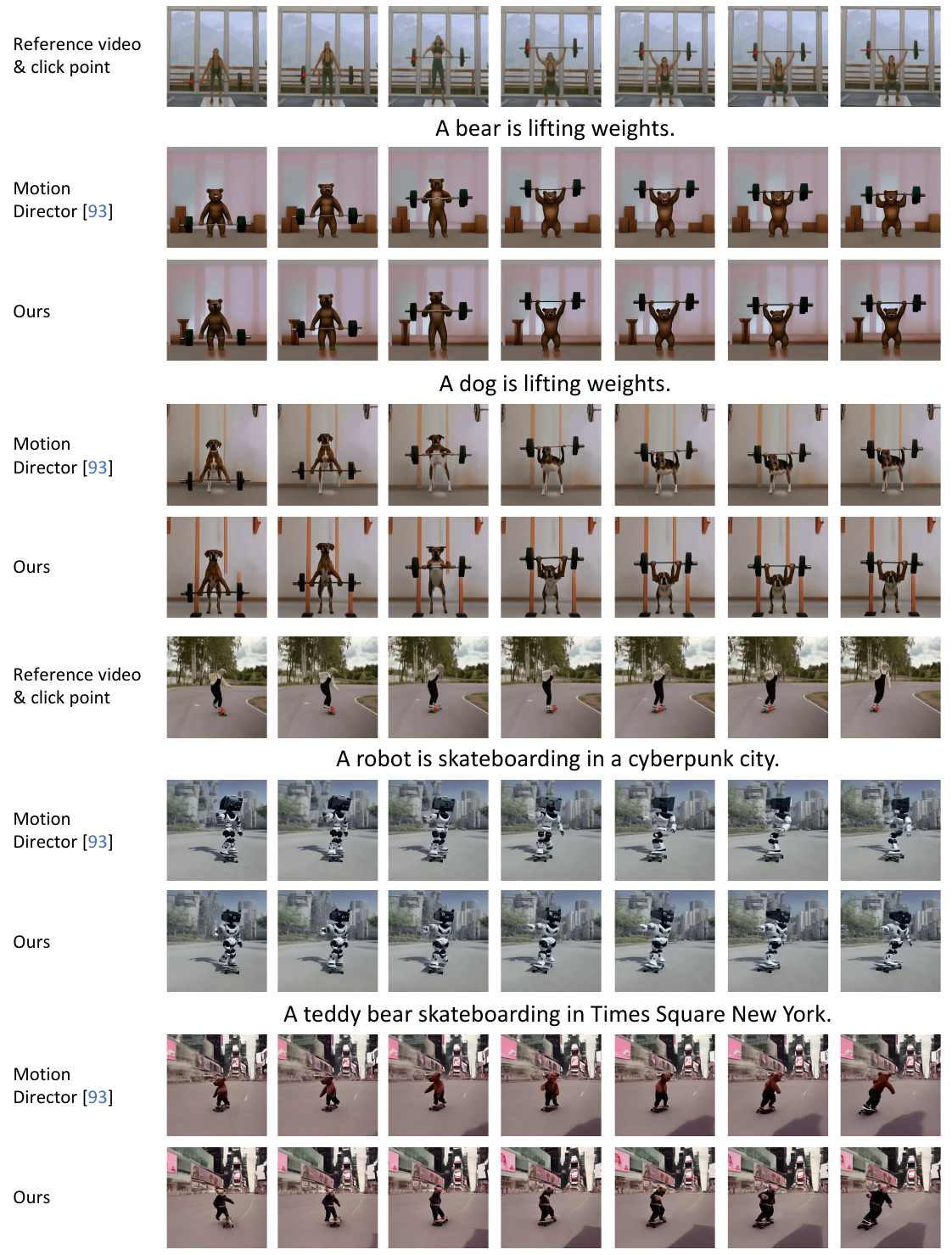}
\caption{\textbf{Qualitative comparison of reference video control.} We evaluate our method and MotionDirector~\cite{zhao2025motiondirector}. To ensure a fair comparison, we use the same initial noises for \cite{qiu2024freetraj} and our method.
Our method demonstrates superior motion alignment with the reference video and improved temporal coherence. (Best viewed with zoom for finer details.)
}
\label{fig:refer_video_compar_supp}
\end{figure*}

We provide more comparison examples of trajectory control based methods in Figure~\ref{fig:traj_compar_supp} and reference video control based methods in Figure~\ref{fig:refer_video_compar_supp}.

\myparagraph{Limitation and failure case.}
Our method focuses on generating videos that simulate the motion of reference videos. However, when the target prompt deviates significantly from the reference video, the results are suboptimal. For instance, a car struggles to follow a ``V'' trajectory, and it is unrealistic to make a dolphin lift weights, as illustrated in Figure~\ref{fig:failure_case_supp}.

